\documentclass[11pt,a4paper]{article}
\usepackage[hyperref]{emnlp2018}
\usepackage{times}
\usepackage{latexsym}

\usepackage{comment}
\usepackage{amsmath}
\usepackage{amsfonts}
\usepackage{graphicx}
\usepackage{subcaption}
\DeclareMathOperator*{\E}{\mathbb{E}}

\usepackage{url}

\usepackage{multirow}
\usepackage{adjustbox}
\usepackage{array}

\newcolumntype{R}[2]{%
    >{\adjustbox{angle=#1,lap=\width-(#2)}\bgroup}%
    l%
    <{\egroup}%
}

\aclfinalcopy 

\title{Proximal Policy Optimization and its Dynamic Version \\ for Sequence Generation}

\author{Yi-Lin Tuan\thanks{$^*$ indicates equal contribution.} \\
  National Taiwan University \\
  {\tt pascaltuan@gmail.com} \\\And
  Jinzhi Zhang$^*$ \\
  HuaZhong University of science and technology \\
  {\tt t06902115@ntu.edu.tw} \\\AND
  Yujia Li \\
  HuaZhong University of science and technology \\
  {\tt 15926479126@163.com} \\\And
  Hung-yi Lee \\
  National Taiwan University \\
  {\tt hungyilee@ntu.edu.tw} \\}

\date{}

\begin{document}
\maketitle
\begin{abstract}
In sequence generation task, many works use policy gradient for model optimization to tackle the intractable backpropagation issue when maximizing the non-differentiable evaluation metrics  or fooling the discriminator in adversarial learning.
In this paper, we replace policy gradient with proximal policy optimization (PPO), which is a proved more efficient reinforcement learning algorithm, and propose a dynamic approach for PPO (PPO-dynamic). 
We demonstrate the efficacy of PPO and PPO-dynamic on conditional sequence generation tasks including synthetic experiment and chit-chat chatbot. 
The results show that PPO and PPO-dynamic can beat policy gradient by stability and performance.
\end{abstract}

\section{Introduction}
The purpose of chit-chat chatbot is to respond like human when talking with people.
Chit-chat has the one-to-many property, that is, given an input sentence, there are many possible answers.
For example, when a user says \emph{``How is the weather?''}, the ideal responses include \emph{``Today's weather is good.''} and \emph{``It's raining.''}, etc.

The recent success of sequence-to-sequence model~\citep{sutskever2014sequence} as chatbot~\citep{vinyals2015neural} inspires researchers to study how to improve generative model based chatbot to beat the rule-based and retrieval-based chatbots by coherence, creativity and its main disadvantage: robustness.
Many state-of-the-art algorithms are thus applied to this text generation task, such as generative adversarial networks (GANs)~\citep{yu2017seqgan,che2017maximum,gulrajani2017improved,lin2017adversarial,tuan2018stepgan} and reinforcement learning~\citep{ranzato2015sequence}.

Particularly, the policy gradient based method REINFORCE~\citep{ranzato2015sequence} is used to optimize the BLEU score~\citep{papineni2002bleu} in text generation, and policy gradient with Monte-Carlo tree search (MCTS) is also used to optimize sequence generative adversarial network (SeqGAN)~\citep{yu2017seqgan,li2017adversarial}.
Despite of the reported good performance, policy gradient empirically leads to destructively updates and thus easily adopts similar actions.
Moreover, policy gradient makes the training of SeqGAN more unstable than regular GANs.
The recent proposed method, proximal policy optimization (PPO)~\citep{schulman2017proximal}, can deal with the problems by regularizing the gradient of policy.

Because we have observed that the unstability of policy gradient in both reinforcement learning and GANs limits the performance, it is desired to replace policy gradient with the more efficient optimization method: PPO. In addition, we modify the constraints of PPO to make it both dynamic and more flexible, and show that this modification can further improve the training.

\section{Related Works}
Previous pure reinforcement learning based approaches attempt to fine-tune text generation model to optimize BLEU scores. 
These approaches include REINFORCE, MIXER~\citep{ranzato2015sequence}, and actor-critic approach~\citep{bahdanau2016actor}.
They are the very first attempts to apply reinforcement learning to seq2seq and show promising performance. 
Nonetheless, on Atari games and continuous control domains, many approaches are proposed to improve the scalability~\citep{schulman2015trust}, data efficiency~\citep{popov2017data}, and robustness~\citep{pinto2017robust}.
These techniques are not yet explored on text generation. 

A recent well-known method for improving  chit-chat chatbot is using adversarial learning (i.e., GANs)~\citep{li2017adversarial, che2017maximum, lin2017adversarial, rajeswar2017adversarial, press2017language, tuan2018stepgan}. Because the discrete text makes the backpropagation through GANs intractable, researchers have proposed several approaches to deal with it. Among them, Gumbel-softmax~\citep{kusner2016gans} and policy gradient~\citep{yu2017seqgan} are the most widely used.

Policy gradient has shown promising results but still left rooms for improvements. By intuition, policy optimization methods that have been proved much more efficient than policy gradient should be applied to the conditional text generation task.

\section{Background}

We use gated recurrent unit (GRU) to build a sequence-to-sequence model (seq2seq) as our chit-chat chatbot. The seq2seq model contains an encoder and a decoder that the encoder reads in an input sentence $\{x_t\}_{t=1}^N$ and the decoder predicts an output sentence $\{y_t\}_{t=1}^M$, where $x_t$ and $y_t$ are words, and $N$ and $M$ are the length of input and output respectively.

Sentence generation can be formulated as a Markov decision process (MDP)$~\sim(S,A,T,R,\gamma)$, where $S$ is a set of states $s_t=\{x,y_{1:t-1}\}$, $A$ is a set of actions $a_t=y_t$, $T$ is the transition probability of the next state given the current state and action\footnote{In sentence generation, transition probability is not needed because given the current state and action, the next state is determined, that is, $T(s_{t+1}=\{x,y_{1:t}\}|s_t=\{x,y_{1:t-1}\},a_t=y_t )=1$.}, $R$ is a reward function $r(s_t,a_t)$ for every intermediate time step $t$, and $\gamma$ is a discount factor that $\gamma \in [0,1]$. 
The actions are taken from a probability distribution called policy $\pi$ given the current state (i.e., $a_t\sim \pi(s_t)$). 
In sentence generation, $\pi$ is a seq2seq model. 
Therefore, reinforcement learning methods are suitable to apply to sentence generative model by learning the seq2seq model, or policy $\pi$, that can gain as much as possible reward.

As previous works, we can use two types of reward functions: (1) task-specific scores (i.e., BLEU~\citep{papineni2002bleu}), (2) discriminator scores in GANs~\citep{li2017adversarial}.

\subsection{Policy Gradient}
Given reward $r_t$ at each time step $t$, the parameter $\theta$ of policy $\pi$ (a seq2seq model) is updated by policy gradient as following: 
\begin{equation}
\nabla_\theta = A_t \nabla_\theta \pi_\theta(a_t,s_t),
\end{equation}
where $A_t=\sum_{\tau=t}^M \gamma^{\tau-t} r_\tau  - b$ is the 1-sample estimated advantage function, in which $b$ is the baseline\footnote{The baseline $b$ is often a value function that equals to the expected obtained rewards at the current state $s_t$.} to reduce the training variance\footnote{The term $\sum_{\tau=t}^M \gamma^{\tau-t} r_\tau$ is a 1-sample estimation of expected obtained rewards at current time step $t$ after generating the whole sentence.
It is the summation of the intermediate rewards from time $t$ to the end, each is multiplied by a discount factor $\gamma$.}.
$A_t$ can be interpreted as the goodness of adopted action $a_t$ over all the possible actions at state $s_t$.
Policy gradient directly updates $\theta$ to increase the probability of $a_t$ given $s_t$ when advantage function is positive, and vise versa.

\subsection{Proximal Policy Optimization}
Proximal policy optimization (PPO)~\citep{schulman2017proximal} is modified from trust region policy optimization (TRPO)~\citep{schulman2015trust}, and both methods aim to maximize a surrogate objective and subject to a constraint on quantity of policy update:
\begin{equation}
\begin{split}
	& \underset{\theta}{\max} L^{TRPO}(\theta), L^{TRPO}(\theta)=\E [\frac{\pi_\theta(a_t|s_t)}{\pi_{\theta_{old}} (a_t|s_t)} A_t], \\
    & \textup{subject to~~} \E [\textup{KL}[\pi_{\theta_{old}} (a_t|s_t),\pi_\theta(a_t|s_t)]] \leq \delta. \\
\end{split}
\end{equation}
$\theta_{old}$ is the old parameters before update.
Because the KL-divergence between $\pi_\theta$ and $\pi_{\theta_{old}}$ is bounded by $\delta$, the updated policy $\pi_\theta$ cannot be too far away from the old policy $\pi_{\theta_{old}}$.

PPO uses a clipped objective to heuristically constrain the KL-divergence:
\begin{equation}
\label{eq:ppo}
\begin{split}
    & \underset{\theta}{\max} L^{PPO}(\theta), \\
    &  L^{PPO}(\theta) = \E [\min(\rho_t A_t, clip(\rho_t,1-\epsilon, 1+\epsilon) A_t)] \\
\end{split}
\end{equation}
where $\rho_t = \frac{\pi_\theta(a_t|s_t)}{\pi_{\theta_{old}} (a_t|s_t)}$ and $\epsilon$ is a hyperparameter (e.g., $\epsilon=0.1$).
When $A_t$ is positive, the objective is clipped by $(1+\epsilon)$; when $A_t$ is negative, the objective is clipped by $(1-\epsilon)$.
$L^{PPO}$ excludes the changes that will improve the objective, and includes the changes that will make the objective worse\footnote{When $A_t$ is positive, the objective is clipped by $(1+\epsilon)$; when $A_t$ is negative, the objective is clipped by $(1-\epsilon)$.}. 

\section{Proposed Approach}

We propose to replace policy gradient on conditional text generation with PPO, which constraints the policy update by an assigned hyperparameter $\epsilon$.
Given the input sentence $\{x_t\}_{t=1}^N$, seq2seq predicts an output sentence $\{y_t\}_{t=1}^M$, where the words $y_t$ are sampled from probability distribution $\pi_\theta(y_t| x,y_{1:t-1})$. By PPO, the update of seq2seq
is to maximize $L^{PPO}$ in Equation~(\ref{eq:ppo}), where $\rho_t = \frac{\pi_\theta(y_t| x,y_{1:t-1})}{\pi_{\theta_{old}}(y_t| x,y_{1:t-1})}$.

In theory, the fixed hyperparameter $\epsilon$ that aims to bound the KL-divergence is not consistent with that KL-divergence is actually depending on the old policy $\pi_{old}$. 
Instead, we propose dynamic parameters that automatically adjust the bound to have better constraints, and call it PPO-dynamic in Section~\ref{sec:exp}. 
The optimization is thus modified as below:
\begin{equation}
\begin{split}
\nabla_\theta  = \nabla_\theta & \min(\rho_t A_t, clip(\rho_t, 1-\beta, 1+\alpha) A_t)\\
\textup{where~} \beta & = \min(\beta_1, \beta_2 \sqrt{1/\pi_{\theta_{old}}-1} )\\
\alpha & = \min(\alpha_1, \alpha_2 \sqrt{1/\pi_{\theta_{old}}-1} )
\end{split}
\end{equation}
where $\beta_1$, $\beta_2$, $\alpha_1$ and $\alpha_2$ are hyperparameters, and the derivation of the term $\sqrt{1/\pi_{\theta_{old}}-1}$ is written in the Supplementary Material. In most cases, it is sufficient to use $\alpha_1 = \beta_1 = \inf$ and $\alpha_2 = \beta_2$, so only one hyperparameter is left to tune. This setup is thus comparable to the original PPO.

We can interpret PPO-dynamic by that it gives bigger gradient tolerances for actions that have lower probability (i.e., $\pi_{\theta_{old}}(y_t|x,y_{1:t-1})$ is small), and vise versa.
This mechanism then dynamically looses and tightens $\epsilon$ of PPO throughout the training.

\section{Experiments}
\label{sec:exp}
To validate the efficacy of using PPO and our proposed PPO-dynamic, we compare them with REINFORCE, MIXER, and  SeqGAN with policy gradient. 

\subsection{Synthetic Experiment: Counting}
{\it Counting}~\citep{tuan2018stepgan} is a one-to-many conditional sequence generation task that aims to count the length and position of input sequence. Each input sequence can be represented as $\{x_t\}_1^N$, where $x_t$ are digits 0-9 
and $N$ is ranging from 1 to 10; each output sequence is a three digits sequence that can be represented as $\{y_t\}_1^M$, where $y_t$ are digits 0-9 and $M$ must be 3. The output sequence must obey the rule that $y_2 = x_t$, where $t$ is randomly selected from $\{1...N\}$, and $y_1 = t-1$ and $y_3 = N-t$.
For example, given an input sequence $\{9, 2, 3\}$, the possible output sequences contains $\{0, 9, 2\}$, $\{1, 2, 1\}$ and $\{2, 3, 0\}$

Because it is easy to judge if a generated sequence is correct or not, we can directly optimize the correctness, and estimate the {\it precision}, which is the number of correct answers divided by the number of all generated answers.

\subsubsection{Results}
\begin{figure}[t!]
\centering
\includegraphics[width=.8\linewidth]{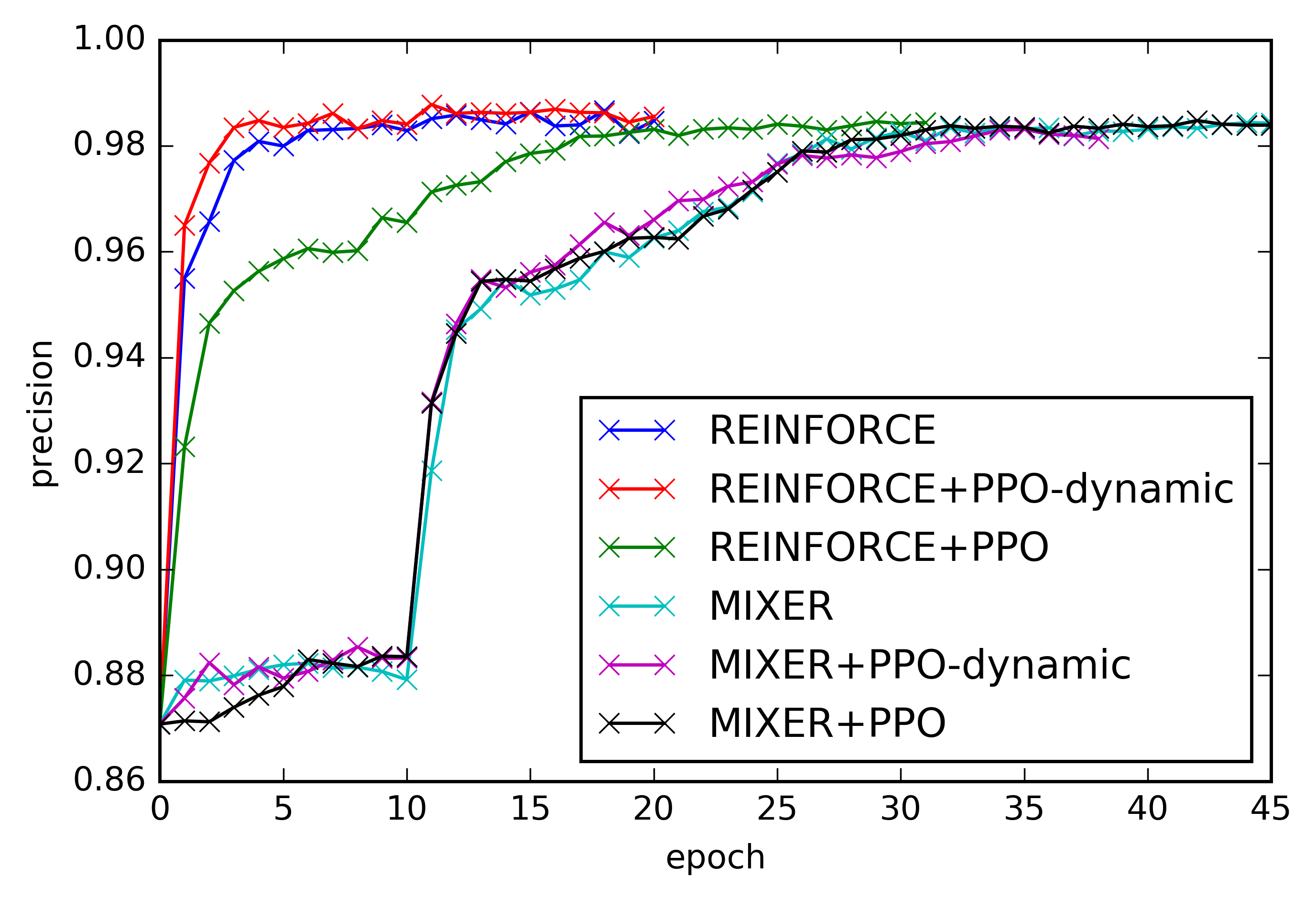}
\caption{Learning curve of different algorithms on synthetic counting task.}
\label{fig:count-reward}
\end{figure}

In Table~\ref{tab:prec} and Figure~\ref{fig:count-reward}, we compares the precision and learning curves of different algorithms on the counting task.
As shown in Table~\ref{tab:prec}, we observed a tremendous improvement in precision of SeqGAN by using PPO-dynamic instead of policy gradient, and PPO-dynamic can also achieve comparable performance (which is a very high precision) with REINFORCE and MIXER.
The learning curves are plotted in Figure~\ref{fig:count-reward}.
We find out that the training progress of PPO is slower than PPO-dynamic, which validates that dynamic constraints can facilitate the training.

\begin{table}[t!]
	\centering
	\begin{tabular}{l|ll}
    \hline
    \multicolumn{2}{l}{Algorithms} & precision \\
    \hline
    \multicolumn{2}{l}{MLE} & 87.01  \\\hline
    \multicolumn{2}{l}{REINFORCE}   & 98.61  \\
    \multicolumn{2}{l}{PPO}         & 98.42  \\
    \multicolumn{2}{l}{PPO-dynamic} & 98.62  \\\hline
    \multirow{3}{*}{MIXER}
    & original 		& 98.43  \\
    & PPO   	 	& 98.39  \\
    & PPO-dynamic   & 98.42  \\\hline
    \multirow{3}{*}{SeqGAN}
    & original    &  87.37  \\
    & PPO   	  &  87.24  \\
    & PPO-dynamic &  91.85  \\

	\hline
	\end{tabular}
    \caption{Results of different algorithms on counting task.}
	\label{tab:prec}
\end{table}

To demonstrate the ability of giving diverse answers, we show the distribution of $y_1$, a number related to input sentence length $N$, given a fixed $N$. 
Figure 2 shows the $y_1$ distribution of three policy optimization methods and the ground truth distribution when fixing $N=5$\footnote{For other sentence lengths, please refer to Supplementary Material.}.
We can see that using REINFORCE will severely make the probability distribution concentrate on one word. On the other hand, the distribution given by PPO and PPO-dynamic are much closer to the ground truth distribution. 

\begin{figure}[t]
\centering
\includegraphics[width=.7\linewidth]{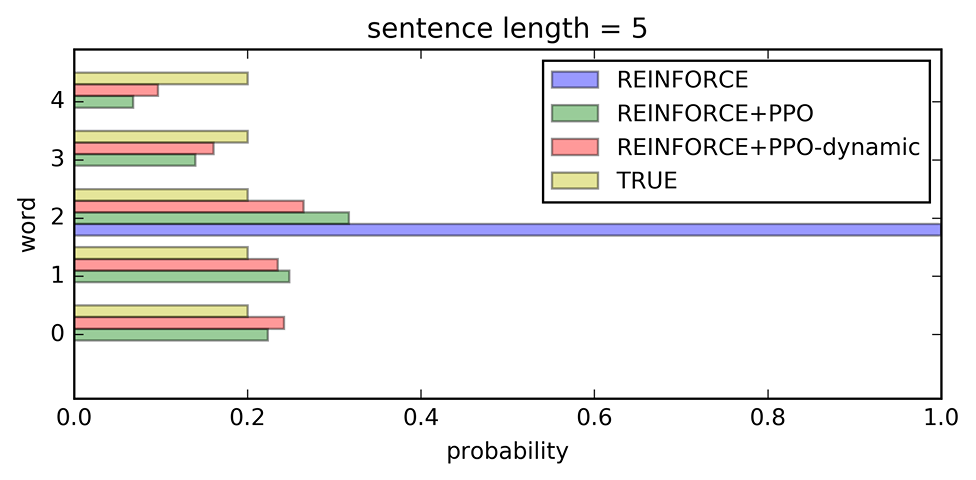}
\caption{The average distribution of the first output $y_1$ when the input sentence length $N$ equals to 5. The possible outputs are ranging from 0 to 4. The yellow bar shows the ground truth probability of each word.}
\label{fig:count-bar-N=5}
\end{figure}

\subsection{Chit-chat Chatbot: OpenSubtitles}
For training chit-chat chatbot, we tested our algorithms on OpenSubtitles dataset~\citep{Tiedemann:RANLP5} and used BLEU-2~\citep{papineni2002bleu,liu2016not}\footnote{Previous work~\citep{liu2016not} suggests to use BLEU-2, which is proved to be more consistent with human score than BLEU-3 and BLEU-4.} as the reward for reinforcement learning. Specifically, we organized both our training and testing data, and made them become that each input sentence has one to many answers. This made the evaluation of BLEU-2 score more correct by having multiple references.


\subsubsection{Results}
\begin{figure}[t!]
\centering
\includegraphics[width=.8\linewidth]{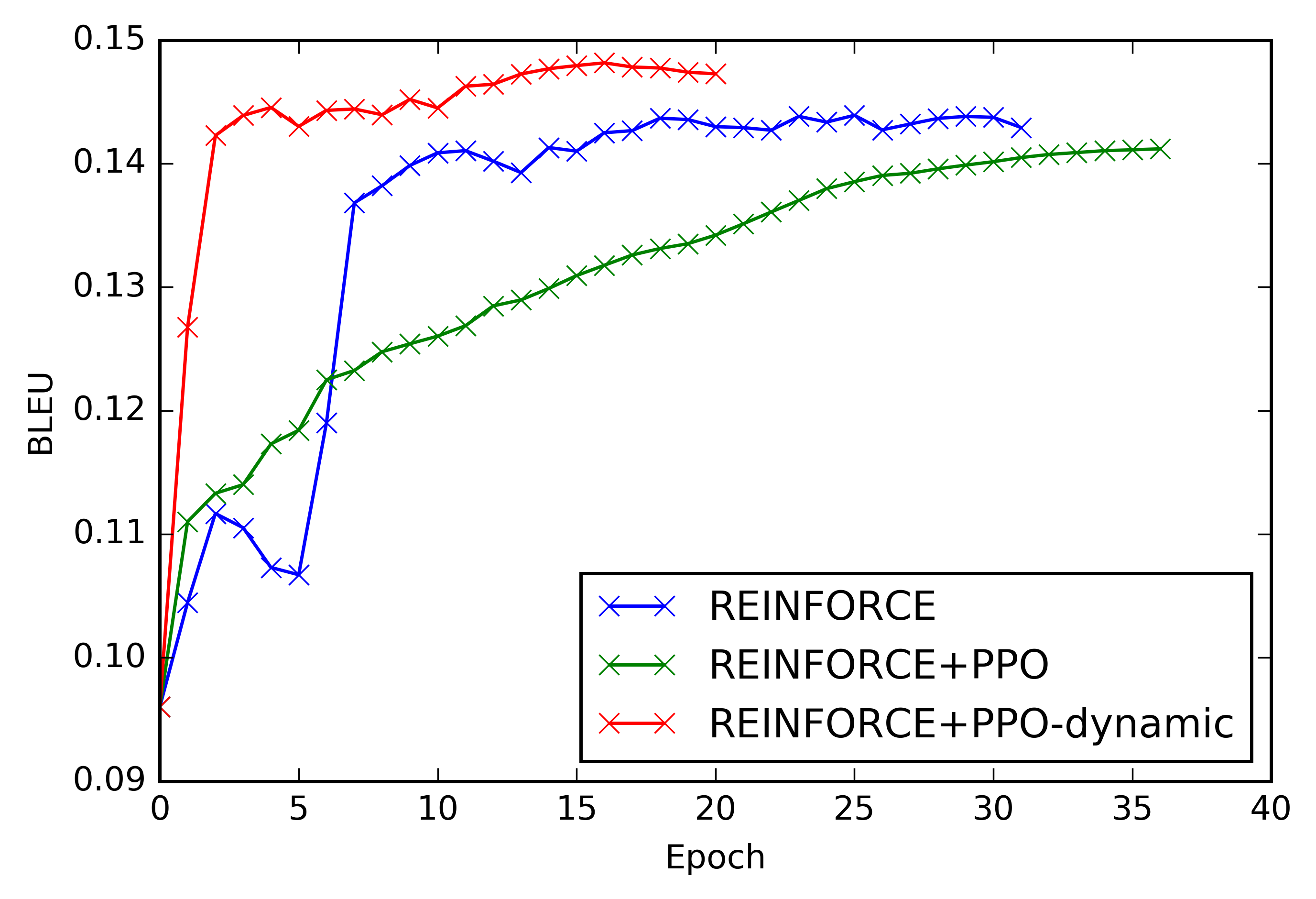}
\caption{Learning curve of BLEU-2 for different algorithms on OpenSubtitles.}
\label{fig:real-reward}
\end{figure}

The BLEU-2 scores and learning curves of different optimization algorithms are presented in Table~\ref{tab:bleu} and Figure~\ref{fig:real-reward}.
From the testing results in Table~\ref{tab:bleu}, we can see that the three optimization methods have comparable performance, but PPO-dynamic achieves a slightly higher BLEU-2 score than REINFORCE and PPO.
Moreover,
we find out that the training progress of both PPO and PPO-dynamic are more stable than policy gradient, and the training progress of PPO-dynamic is much faster than the others.
This shows that PPO based methods can improve the high variance problem of using REINFORCE, and the dynamic constraint can help the learning converge quickly.

We demonstrate some responses of candidate model in Supplementary Material.

\begin{table}[t!]
	\centering
	\begin{tabular}{l|ll}
    \hline
    \multicolumn{2}{l}{Algorithms} & BLEU \\
    \hline
    \multicolumn{2}{l}{MLE} & 9.62  \\\hline
    \multicolumn{2}{l}{REINFORCE}   &  14.29 \\
    \multicolumn{2}{l}{PPO}         & 14.12  \\
    \multicolumn{2}{l}{PPO-dynamic} & 14.73  \\\hline
    
	\hline
	\end{tabular}
    \caption{The BLEU-2 Results of different algorithms on OpenSubtitles.}
	\label{tab:bleu}
\end{table}

\section{Conclusion}
In this paper, we replace policy gradient with PPO on conditional sequence generation, and propose a dynamic approach for PPO to further improve the optimization process. 
By experiments on synthetic task and chit-chat chatbot, we demonstrate that both PPO and PPO-dynamic can stabilize the   training and lead the model to learn to generate more diverse outputs. 
In addition, PPO-dynamic can speed up the convergence. 
The results suggest that PPO is a better way for sequence learning, and GAN-based sequence learning can use PPO as the new optimization method for better performance.


\bibliography{emnlp2018}
\bibliographystyle{acl_natbib_nourl}

\end{document}


\maketitle

\begin{abstract}
  This supplementary material contains the derivation of our proposed PPO-dynamic, the experimental settings and more results.
\end{abstract}

\appendix

\section{Derivation of the Constraints of PPO-dynamic}
Because we want to find real constraints for PPO, we define that $\frac{P(a)}{P_{old}(a)} = 1+\alpha(a)$, and aim to find $\alpha(a)$ so that $KL(P_{old}||P) \leq \delta$.
First,

\begin{equation}
\begin{split}
	& KL(P_{old}||P) = -\int_{x} P_{old}(x)\ln\frac{P(x)}{P_{old}(x)} dx\\
    & = -P_{old}(a)\ln\frac{P(a)}{P_{old}(a)} + \int_{x\neq a} P_{old}(x)\ln\frac{P_{old}(x)}{P(x)} dx\\
    & = -P_{old}(a)\ln(1+\alpha(a)) + \int_{x\neq a} P_{old}(x)\ln\beta(x) dx
\end{split}
\end{equation}
where $\beta(x)$ is defined as $\beta(x) = \frac{P_{old}(x)}{P(x)}$.

By assuming $\beta(x)$ is a constant $\beta$ for all $x\neq a$, we can find the value of $\beta$ by following:
\begin{equation}
\begin{split}
	&1 = \int_{x\neq a} P(x)dx + P(a)\\
    &= \frac{1}{\beta}\int_{x\neq a} P_{old}(x)dx + (1+\alpha(a))P_{old}(a)\\
   &=\frac{1}{\beta}(1 - P_{old}(a)) + (1+\alpha(a))P_{old}(a)
\end{split}
\end{equation}
so,
\begin{equation}
\begin{split}
    &\beta = \frac{1 - P_{old}(a)}{1 - (1 + \alpha(a))P_{old}(a)}
\end{split}
\end{equation}

After substituting the term $\beta(x)$ in Equation~(1) by Equation~(3), we can get:

\begin{equation}
\begin{split}
     &KL(P_{old}||P) = -P_{old}(a)\ln(1 + \alpha(a)) \\
     &+ \big(1 - P_{old}(a)\big)\ln\Big(\frac{1 - P_{old}(a)}{1 - (1 + \alpha(a))P_{old}(a)}\Big)\\
     & = -P_{old}(a)\ln(1 + \alpha(a)) \\
     & - \big(1 - P_{old}(a)\big)\ln\Big(1 - \alpha(a)\frac{P_{old}(a)}{1 - P_{old}(a)}\Big)\\
\end{split}
\end{equation}

Now we assume that $\alpha(a) \ll 1$, and then we can use Taylor expansion for $\ln(1 + \alpha(a))$ and $\ln(1 - \alpha(a)\frac{P_{old}(a)}{1 - P_{old}(a)})$ such that:

\begin{equation}
\begin{split}
	& \ln(1 + \alpha(a)) \doteq \alpha(a) - \frac{\alpha(a) ^{2}}{2} ,\\
    & \ln(1 - \alpha(a)\frac{P_{old}(a)}{1 - P_{old}(a)})\\
	&\doteq -\alpha(a)\frac{P_{old}(a)}{1 - P_{old}(a)} - \frac{\alpha(a)^{2}}{2}\frac{P_{old}(a)^{2}}{(1 - P_{old}(a))^{2}}
\end{split}
\end{equation}

By substituting the terms in Equation~(4), we have:

\begin{equation}
\begin{split}
     &KL(P_{old}||P) \doteq -P_{old}(a)(\alpha(a) - \frac{\alpha(a) ^{2}}{2}) \\
     & + (1 - P_{old}(a))\Big(-\alpha(a)\frac{P_{old}(a)}{1 - P_{old}(a)}\\
     & - \frac{\alpha(a)^{2}}{2}\frac{P_{old}(a)^{2}}{(1 - P_{old}(a))^{2}}\Big)\\
     & = \frac{P_{old}(a)}{1 - P_{old}(a)} \frac{\alpha(a)^{2}}{2} \leq \delta
\end{split}
\end{equation}

so we get,
\begin{equation}
\begin{split}
     & - \sqrt{2\delta}\sqrt{\frac{1 - P_{old}(a)}{P_{old}(a)}}\leq \alpha(a)\leq\sqrt{2\delta}\sqrt{\frac{1 - P_{old}(a)}{P_{old}(a)}}
\end{split}
\end{equation}
that is, when we want to constrain PPO by restricting $\frac{P(a)}{P_{old}(a)}$, we have to constrain $\alpha(a)$ by $\sqrt{\frac{1 - P_{old}(a)}{P_{old}(a)}}$, or $\sqrt{\frac{1}{P_{old}(a)}-1}$.

\begin{figure*}[t!]
\centering
\includegraphics[width=1.0\textwidth]{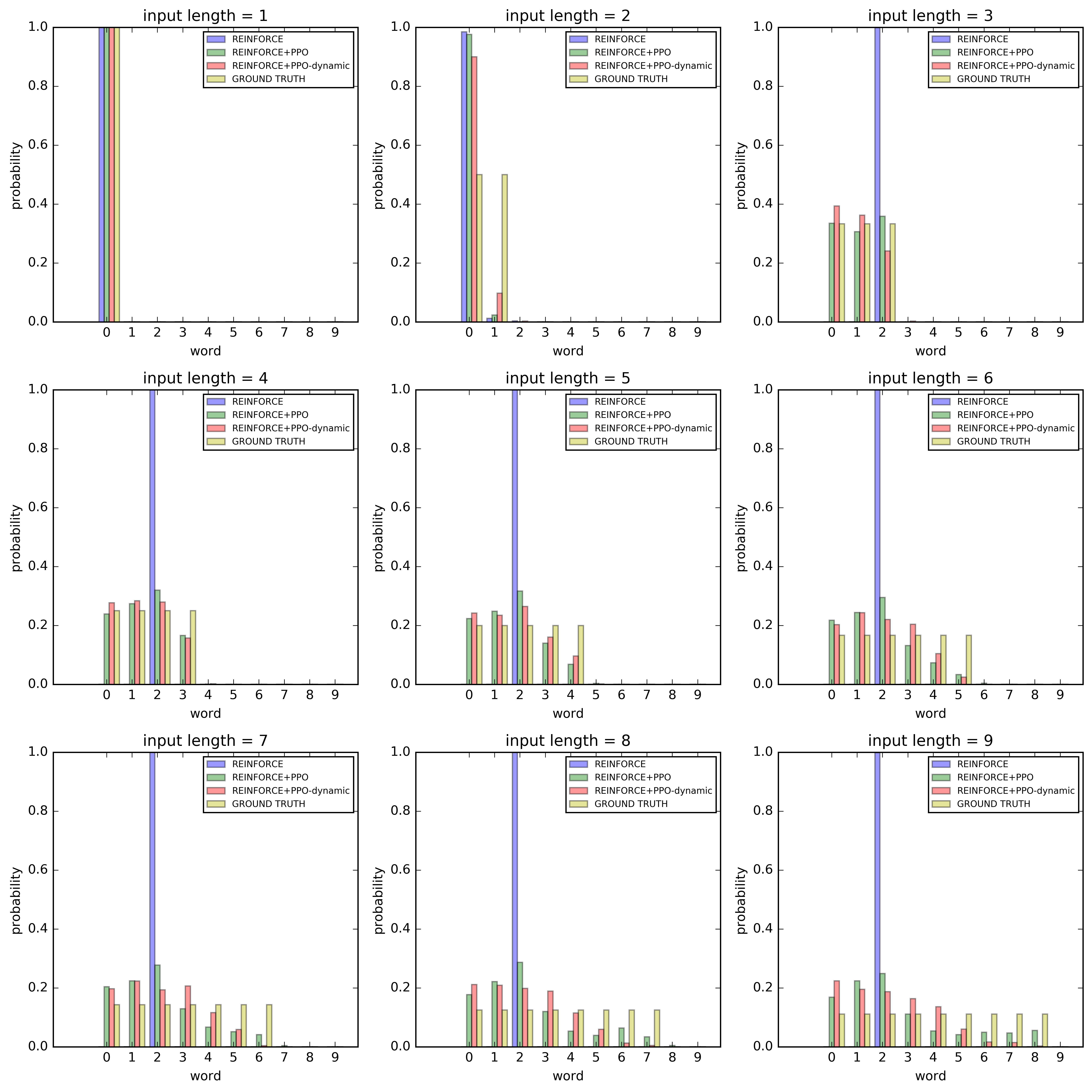}
\caption{The average distribution of the first output with different input sentence length. We plot the probability from 0 to 9. The yellow box is the ground truth probability of each word. The blue box and the red box shows the probability of each word trained by REINFORCE and REINFORCE with dynamic PPO.
}
\label{fig:count-all-bar}
\end{figure*}

\section{Pseudo Code of PPO and PPO-dynamic}
The pseudo code is listed in Algorithm 1 and is basically the same as PPO algorithm~\citep{schulman2017proximal}.

\begin{table}[t]
\begin{center}
\begin{tabular}{l}
\hline
\textbf{Algorithm 1} PPO and PPO-dynamic\\\hline
\hspace{0cm} Initialize $\pi$\\
\hspace{0cm} Pretrain $\pi$ using MLE and set $\pi_{old}=\pi$\\
\hspace{0cm} Initialize discriminator $D$, value network $V$\\

\hspace{0cm} {\bf for} number of training iterations {\bf do}\\
\hspace{0.5cm}   sample a batch of data $\{y_t\}_{t=1}^M$ using $\pi_{old}$\\
\hspace{0.5cm}   set $\pi_{old} = \pi$\\
\hspace{0.5cm}   optimize $\pi$ using $L^{PPO}$ and the sampled \\\hspace{0.5cm}   data $\{y_t\}_{t=1}^M$\\
\hspace{0cm} {\bf end for}\\
\hline
\end{tabular}
\end{center}
\end{table}

\section{Experimental Setup}
The baseline in advantage function $A_t$ is trained as \citep{ranzato2015sequence}.
The best hyperparameters we found out by grid search is that,
for REINFORCE and MIXER with PPO-dynamic, $\alpha_1 = \beta_1 =+inf$ and $\alpha_2 = \beta_2 = 1.0$; for SeqGAN with PPO-dynamic, $\alpha_1 = 10.0$, $\beta_1 =0.5$, and $\alpha_2 = \beta_2 = 0.2$; for REINFORCE, MIXER and SeqGAN with original PPO, $\epsilon = 0.2$. 

We use 1 layer GRU with 128 dimension for Counting task, and 1 layer GRU with 512 dimension for chit-chat chatbot.

\section{Distribution of the first output with different input sentence length}
We want to check the distribution of the first output on counting task in order to get an intuitive impression of the ability to response different answers, so we plot the distribution of all the input length in Figure~\ref{fig:count-all-bar}. We can see that using REINFORCE method will make the distribution tremendously concentrate on word 2 when the input sentence length is bigger than 2. On the other hand, using PPO method will generate a more scattered distribution and is much closer to the ground-truth distribution.

\begin{table}[t]
	\centering
	\begin{tabular}{ccc}
   
    \hline
    Length  & REINFORCE & PPO-dynamic \\
    \hline
    1          &0.066 & 0.090      \\
    2        &0.064 & 0.076       \\
    3       & 0.066 & 0.029      \\
    4       & 0.066& 0.018       \\
    5       &  0.066  & 0.014      \\
    6       &0.066& 0.012  \\
    7       &0.066& 0.011 \\
    8       &0.066& 0.011\\
    9      &0.066& 0.010\\
	\hline
	\end{tabular}
    \caption{The mean variance of the distribution of the first generated word. Length means the input sentence length, and we average all the variance of the distribution.}
	\label{tab:variance}
\end{table}
We also evaluate the variance of output distribution. The higher the variance, the sharper the distribution. Therefore, it is intuitive to claim that a shorter length input should correspond to a higher variance, and vise versa. We can clearly see the claim in Table~\ref{tab:variance}. REINFORCE always generate a sharp distribution; PPO-dynamic can adjust the distribution according to the input.

\section{Some output example}
In Table~\ref{tab:example}, we present some examples of our trained chit-chat chatbot by optimizing BLEU-2 scores using REINFORCE, PPO and PPO-dynamic.

\begin{table*}[t!]
\centering
\begin{tabular}{r|l}
\hline
input & what can i do for you.\\\hline
MLE & i 'm gonna go back .\\
REINFORCE & go !\\
PPO &  i don 't know what i 'm talking about .\\
PPO-dynamic & i don 't know .\\
\hline\hline
input &  go , go , go !\\\hline
MLE &  go , go , go , go , go !\\
REINFORCE & go !\\
PPO & go on the way .\\
PPO-dynamic & go ! go get him go !\\
\hline\hline
input & i said ...\\\hline
MLE &  i 'm sorry .\\
REINFORCE &  i 'm not .\\
PPO &  i 'm sorry .\\
PPO-dynamic & oh , i 'm sorry .\\
\hline\hline
input & it 's wonderful .\\\hline
MLE &  i 'm not gonna be there .\\
REINFORCE & it 's you .\\
PPO & it 's a good idea .\\
PPO-dynamic & it 's a good idea .\\
\hline\hline
input & i don 't know what you 're talking about .\\\hline
MLE & you 're gonna be a man .\\
REINFORCE & i 'm not .\\
PPO & you 're not gonna be a good man who 's not really you think .\\
PPO-dynamic & you 're gonna be a good man , you know .\\
\hline\hline
input & come here\\\hline
MLE & i 'll be back in the morning .\\
REINFORCE & come here .\\
PPO & i 'm not you 're a little girl .\\
PPO-dynamic & you 're not my father .\\
\hline\hline
input & get him down .\\\hline
MLE & get him out of here !\\
REINFORCE & get him ...\\
PPO & get him out of it .\\
PPO-dynamic & get him out of there .\\
\hline\hline
input & change your mind .\\\hline
MLE & i 'll be back in the morning .\\
REINFORCE & i 'll not .\\
PPO & i 'm not you sure i can 't do it .\\
PPO-dynamic & i 'm not my father .\\
\hline\hline
input & let him go .\\\hline
MLE & he 's gonna be there.\\
REINFORCE & he 's a little girl .\\
PPO & he 's a little girl .\\
PPO-dynamic & go on the phone .\\
\hline
\end{tabular}
\caption{Results of different algorithms on real task.}
\label{tab:example}
\end{table*}

\bibliography{supplementary}
\bibliographystyle{acl_natbib_nourl}